\documentclass[letterpaper]{article} %
\usepackage[preprint]{aaai2027}  %
\usepackage[hyphens]{url}  %
\usepackage{graphicx} %
\usepackage{natbib}  %
\usepackage{caption} %
\usepackage{booktabs} %
\usepackage[table]{xcolor}  %
\usepackage{tikz}
\usetikzlibrary{arrows.meta, positioning, calc}
\usepackage{pgfplots}
\pgfplotsset{compat=1.18}
\usepackage{tcolorbox}  %
\providecommand{\tcbline}{\par\vspace{1.5pt}{\color{black!30}\hrule height 0.3pt}\vspace{2.5pt}}
\usepackage{fontawesome5}  %

\newcommand{\MerschReclass}{24.9}   %
\newcommand{\MerschVUS}{7.7}        %
\newcommand{\MakhnoonGC}{40}       %
\newcommand{\MakhnoonOnc}{12.5}    %

\newcommand{\CorpusPosts}{3{,}900{,}607}
\newcommand{\CorpusPostsApprox}{3.9 million}  %
\newcommand{\CorpusSubs}{32}
\newcommand{\CorpusYearStart}{2008}
\newcommand{\CorpusYearEnd}{2026}

\newcommand{\TriageModel}{\texttt{gpt-4o-mini}}

\newcommand{\TotalTriaged}{3{,}900{,}607}

\newcommand{\BTwoCount}{11{,}253}
\newcommand{\BTwoPct}{0.288}

\newcommand{\RateSpread}{850}           %

\newcommand{\SpecRate}{6.30}        %
\newcommand{\SpecCorpusPct}{2.3}    %
\newcommand{\SpecBTwoShare}{50.1}   %
\newcommand{\GenRate}{0.10}         %
\newcommand{\GenCorpusPct}{75.4}    %

\newcommand{\CorpusAuthorsApprox}{1.4 million}  %
\newcommand{\CorpusReplyPct}{90.8}       %
\newcommand{\BTwoAuthors}{9{,}395}
\newcommand{\BTwoReplied}{94}

\newcommand{\CostPerPost}{0.000178}     %
\newcommand{\CostCleanTotalUSD}{698}    %

\newcommand{\CtrlSetTotal}{6,940}
\newcommand{\CtrlBTwoSet}{4,500}
\newcommand{\CtrlBOneSet}{1,361}
\newcommand{\CtrlIrrSet}{1,079}
\newcommand{\CtrlNegTotal}{2,440}
\newcommand{\CtrlBTwoRecall}{98.1}
\newcommand{\CtrlBTwoPrec}{99.95}
\newcommand{\CtrlBTwoOK}{4,415}
\newcommand{\CtrlBTwoTrue}{4,500}
\newcommand{\CtrlBTwoRouted}{4,417}
\newcommand{\CtrlOverCalls}{2}

\newcommand{\CtrlReviewBTwo}{100}

\newcommand{\BTwoNotActionable}{9{,}383}
\newcommand{\BTwoNotResolvable}{1{,}471}
\newcommand{\BTwoResolvedCV}{399}

\newcommand{\BTwoUnchanged}{237}
\newcommand{\BTwoOutreach}{123}

\newcommand{\OutToConflicting}{112}
\newcommand{\OutToPathogenic}{5}

\newcommand{\SpikeControlN}{1,000}

\newcommand{\SpikeExtPct}{97.0}

\newcommand{\SpikeResPct}{99.6}
\newcommand{\SpikeGeneOnlyN}{334}

\newcommand{\SOneExtCorrect}{2,847}
\newcommand{\SOneExtTotal}{2,942}
\newcommand{\SOneExtPct}{96.8}
\newcommand{\SOneResCorrect}{2,314}
\newcommand{\SOneResTotal}{2,322}
\newcommand{\SOneResPct}{99.7}
\newcommand{\SOneGeneOnlyN}{1,473}

\newcommand{\OutreachDevN}{30}
\newcommand{\OutreachDevStart}{83.3}
\newcommand{\OutreachDevEnd}{100}
\newcommand{\OutreachNVersions}{6}
\newcommand{\OutreachHeldoutN}{93}
\newcommand{\OutreachClinicianN}{30}

\newcommand{\StepThreeAgreePct}{90}
\newcommand{\StepThreeAgreeStartPct}{63}
\newcommand{\StepThreeAgreeCILo}{74}
\newcommand{\StepThreeAgreeCIHi}{97}
\newcommand{\StepThreeSplitN}{3}

\newcommand{\StepThreeSplitDim}{contextualization}

\newcommand{\StepFourFitN}{15}
\newcommand{\StepFourFitTotal}{30}
\newcommand{\StepFourFitPct}{50}

\newcommand{\StepFourAllSixN}{16}

\newcommand{\StepFourAllSixUnfitN}{1}
\newcommand{\StepFourFactualN}{27}
\newcommand{\StepFourFactualUnfitN}{12}

\newif\ifreleased \releasedfalse
\newcommand{\ReleaseVerb}{\ifreleased we release\else upon publication, we will release\fi}

\newcommand{\PilotBTwoTP}{82}  %
\newcommand{\PilotBTwoPredN}{98}  %
\newcommand{\PilotBTwoPrecision}{83.7}  %
\newcommand{\PilotSpecTP}{81}  %
\newcommand{\PilotSpecPredN}{92}  %
\newcommand{\PilotSpecPrecision}{88}  %
\newcommand{\RecallSampleN}{1{,}000}  %
\newcommand{\RecallExpectedBTwo}{3}  %
\newcommand{\RecallCILo}{29}  %
\newcommand{\AblSpecRate}{7.2}  %
\newcommand{\AblGenAdvRate}{0.11}  %
\newcommand{\ArchReleases}{59}  %
\newcommand{\ArchDated}{45}  %
\newcommand{\ArchDatedPct}{37}  %
\newcommand{\ArchUnchanged}{78}  %
\newcommand{\ArchUnchangedPct}{63}  %
\newcommand{\ArchConflUnchanged}{74}  %
\newcommand{\EOneClaudePrecision}{93.5}  %
\newcommand{\EOneCodexPrecision}{94.3}  %
\newcommand{\EOneClaudeNowN}{84}  %
\newcommand{\EOneClaudeNowAgree}{69}  %
\newcommand{\EOnePrismNowOnClaude}{83}  %
\newcommand{\EOneCodexNowN}{74}  %
\newcommand{\EOneCodexNowAgree}{53}  %
\newcommand{\EOneCodexNowAbstain}{17}  %
\newcommand{\EOnePrismNowOnCodex}{73}  %
\newcommand{\EOneClaudeThenN}{79}  %
\newcommand{\EOneClaudeThenAgree}{49}  %
\newcommand{\EOnePrismThenOnClaude}{18}  %
\newcommand{\EOneClaudeCost}{0.19}  %
\newcommand{\EOneCodexCost}{0.31}  %
\newcommand{\PrismCostPerBTwoPost}{0.0005}  %
\newcommand{\EOneCostRatioLo}{380}  %
\newcommand{\EOneCostRatioHi}{620}  %
\newcommand{\OutreachDevStartN}{25}  %
\newcommand{\OutreachDevEndN}{30}  %
\newcommand{\ETwoN}{93}  %
\newcommand{\ETwoTemplateFit}{6}  %
\newcommand{\ETwoReportedFit}{49}  %
\newcommand{\ETwoNoCoachFit}{1}  %
\newcommand{\ETwoCoachFit}{3}  %
\newcommand{\ETwoCoachP}{0.63}  %
\newcommand{\ETwoJudgeKappa}{0}  %
\newcommand{\MonthlyPostsApprox}{40{,}000}  %
\newcommand{\MonthlyRunUSD}{7}  %
\newcommand{\FullRerunUSD}{700}  %
\newcommand{\HarmlessN}{3}  %
\newcommand{\HarmlessFlaggedN}{2}  %
\newcommand{\StepFourVusPlpN}{3}  %
\newcommand{\StepFourVusPlpFitN}{1}  %
\newcommand{\StepFourVusPlpNoGeneN}{2}  %

\newcommand{\PilotN}{300}
\newcommand{\PilotAgreeN}{268}

\newcommand{\PilotKappa}{0.83}

\title{Reclassification Is All You Missed: Surfacing Variant Reinterpretation Need in Online Health Communities at Population Scale}

\author{
    Kamalakkannan Ravi\textsuperscript{\rm 1},
    Mengshu Nie\textsuperscript{\rm 1},
    Annapurna Poduri\textsuperscript{\rm 2},
    Cong Liu\textsuperscript{\rm 1,3}
}
\affiliations{
    \textsuperscript{\rm 1}Division of Genetics and Genomics, Boston Children's Hospital, Harvard Medical School\\
    \textsuperscript{\rm 2}Department of Neurology, Boston Children's Hospital, Harvard Medical School\\
    \textsuperscript{\rm 3}Computational Health Informatics Program, Boston Children's Hospital, Harvard Medical School\\
    Boston, MA, USA\\
    \{kamalakkannan.ravi, adelina.nie, annapurna.poduri, cong.liu\}@childrens.harvard.edu
}

\makeatletter
\g@addto@macro\@maketitle{%
  \vspace{0.5ex}\par
  \noindent\includegraphics[width=\textwidth]{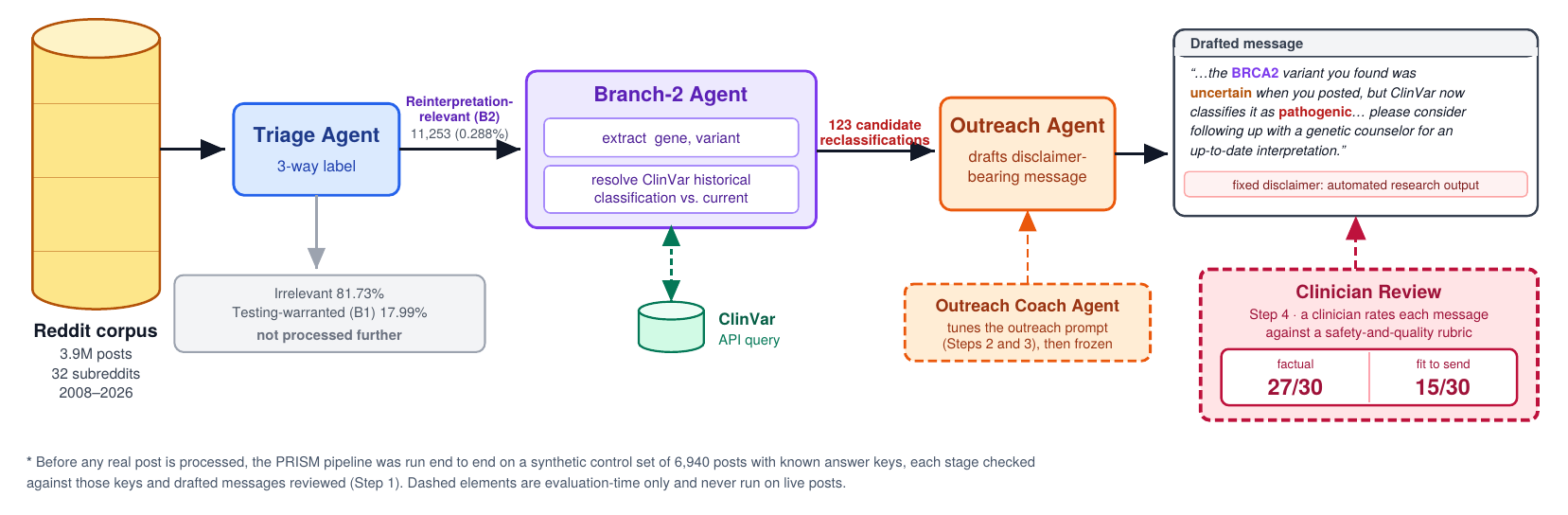}%
  \captionof{figure}{\textbf{Overview of PRISM.} A three-agent cascade: triage labels each of
  \CorpusPostsApprox{} posts, the \BTwoCount{} reinterpretation-relevant (B2) posts advance to
  Branch-2 (which resolves the variant's ClinVar classification as of the test date against the
  current one), and the \BTwoOutreach{} candidate classification differences reach the outreach agent, which
  drafts a disclaimer-bearing outreach message. A collaborating clinician rates each of
  \OutreachClinicianN{} held-out drafts (Step~4); dashed elements run only at evaluation time.}%
  \label{fig:overview}%
  \vspace{1ex}%
}
\makeatother

\usepackage{pdfpages}
\begin{document}
\maketitle

\begin{abstract}
Genetic variants are routinely reclassified in ClinVar after a patient is tested, sometimes from
uncertain to pathogenic, yet patient recontact is inconsistent, with no clear duty and responsibility
diffused among labs, providers, and patients. Many affected people discuss their results in public
online health communities that clinical recontact often does not reach. We built PRISM, an agentic
large-language-model pipeline that screens these posts, checks whether the variant was later
reclassified, and drafts a disclaimer-bearing outreach message, in a no-contact dry run.
Across \CorpusPostsApprox{} posts from \CorpusSubs{} communities,
agent-predicted reinterpretation-relevant discussion is rare (\BTwoPct\%) but concentrated in
specialist genetics communities (\SpecRate\% versus \GenRate\% in general medical-advice communities) and growing fivefold since 2014; PRISM
surfaces \BTwoOutreach{} candidate classification differences, \OutToPathogenic{} of them from an uncertain
submission to pathogenic or likely pathogenic. On a held-out set of \OutreachClinicianN{} real-post drafts, a clinician
judged \StepFourFactualN{} factually grounded and \StepFourFitN{} fit to send:
factual accuracy is necessary but not sufficient for clinically acceptable outreach. PRISM contributes a population-scale reinterpretation-need map
with \ifreleased public code and data\else code and data to be released upon publication\fi.
\end{abstract}

\section{Introduction}
Genetic test results are not static. A result first reported as a \emph{variant of
uncertain significance} (VUS), where the variant is observed but its pathogenicity is
not yet established, is periodically reclassified as evidence accumulates in curated
public archives such as ClinVar and ClinGen, which aggregate variant classifications from clinical
laboratories and expert panels \citep{landrum2018clinvar,clingen2025resource}, and
the revised classification can change clinical management. In one large hereditary-cancer
testing cohort, \MerschVUS\% of unique variants of uncertain significance were reclassified,
rising to \MerschReclass\% of all reported such variants because most recur across individuals
\citep{mersch2018reclass}.

Patients are not systematically informed when a reclassification occurs: an updated report may reach the ordering clinician but often not the patient.
Whether a duty to recontact even exists is unsettled \citep{appelbaum2020duty}, and ACMG guidance
recommends an updated report but assigns no clear responsibility
\citep{deignan2019points,david2019recontact}, so in practice recontact is inconsistent: in a national
survey, only \MakhnoonGC\% of cancer genetic counselors and \MakhnoonOnc\% of oncologists reported
recontacting patients often \citep{makhnoon2023recontact}.

As a result, some patients hold results whose interpretation has changed without their
knowledge, even when the change would alter clinical management, such as added screening or
testing of relatives. A clinic can recontact only patients who remain within its reach, so the
gap falls hardest on those who have changed providers or lack ongoing care. This makes
recontact as much a question of equitable access as of policy; misclassification and reclassification
fall disproportionately on ancestry groups underrepresented in reference data \citep{manrai2016misdiagnoses}.

People discuss prior genetic results in public online health communities, for
example asking whether a VUS reported years earlier has been updated. User-authored
social media is an established, if noisy, source of recoverable clinical signal
\citep{yamaguchi2020histories,wadhwa2023redhot,sehgal2026semaglutide}, and for genetic
results it is a large untapped record: the health communities we assemble
here span \CorpusPostsApprox{} public posts across \CorpusSubs{} communities. Yet the
prevalence and distribution of genetic-variant reinterpretation discussion across them
have never been measured.

This paper introduces PRISM (a name, not an acronym; Figure~\ref{fig:overview}), a pipeline that finds where
reinterpretation need surfaces in public posts, and where it concentrates. We define
\emph{reinterpretation need} as an observable online signal: a post describing a prior genetic
result for which an updated interpretation may be relevant. It is not a patient-level determination
that reinterpretation or recontact is required, which would require consented review of the person's
testing, history, and care. For a post that
references a prior result, PRISM checks whether the classification has since changed. When it detects one,
it drafts a short, disclaimer-bearing message that flags the update and
points to a genetic counselor. Every stage runs in dry-run, and no
message is sent. Our scope is the pipeline itself, not recontact: we design it, validate it on
a control set, run it on real posts, and check whether its drafts are fit to send. We make
four contributions:
\begin{itemize}
\setlength{\itemsep}{2pt}
\item \textbf{A population-scale map of agent-predicted reinterpretation need.} The first characterization of
where it surfaces online, across \CorpusPostsApprox{} posts and \CorpusSubs{} communities:
rare overall, but concentrated in specialist genetics communities (\SpecRate\% versus
\GenRate\% in general medical-advice communities) and growing roughly fivefold since 2014. This
need surfaces in public, where clinical recontact often does not reach.
\item \textbf{A low-cost, reproducible agentic pipeline.} It classifies the full corpus at
\$\CostPerPost{} per post, limits external ClinVar lookups to the small reinterpretation-relevant subset, and drafts
messages from a prompt refined by an \emph{Outreach Coach} agent from reviewer feedback and then
expanded by hand; the reported results use the hand-expanded prompt.
\item \textbf{A responsible, end-to-end evaluation} for the social-media-to-clinical-action
setting: grading on a control set with known answers, plus clinician review of real-post drafts,
all without contacting any user. This evaluation finds factual accuracy necessary but not sufficient for clinically acceptable outreach.
\item \textbf{A \ifreleased public\else planned public\fi{} release consistent with Reddit's Terms of Use.} Following the same practice
as prior Reddit studies \citep{he2026geopolitical,sehgal2026semaglutide}, we do not redistribute raw
post text; \ReleaseVerb{} the code, the full synthetic control set with answer keys, the triage-label and
variant-resolution distributions de-linked from post identifiers, and the generated messages with
fit-to-send labels.
\end{itemize}

\section{Related Work}
\paragraph{The recontact gap.}
Reclassification is common in tested cohorts \citep{mersch2018reclass}, yet recontact is
inconsistent \citep{makhnoon2023recontact}; when it does happen it falls to clinic genetic counselors,
typically by mailed letter, and so reaches only patients still within a clinic's reach \citep{muir2022reclass}. Patient registries such as GenomeConnect deliver
updates directly, but only to those who opt in \citep{savatt2021recontact}. The literature looks for
solutions inside the clinic, not outside, where affected patients already discuss their results in
public.

\paragraph{Mining social media for health signal.}
Those discussions are a resource. Prior work reconstructs rare-disease histories from posts
\citep{yamaguchi2020histories}, turns forum text into annotated datasets such as RedHOT and SMHD
\citep{wadhwa2023redhot,cohan2018smhd,berkane2025webdata}, and recovers population-level signal at scale, from
self-reported drug side effects \citep{sehgal2026semaglutide} to patient experience
\citep{walker2026lupus}; triage pipelines route posts by acuity \citep{hasan2022covid}, with fine-tuned encoders the standard for health-mention classification \citep{naseem2024hmc}. But most of
this aggregates posts into surveillance signal rather than acting on any one author
\citep{paul2016surveillance,freifeld2008healthmap}, and even individual-level studies screen for risk
on consented cohorts \citep{dou2021homeless}. The validity of proxy labels derived this way is itself
debated \citep{chancellor2020methods}.

\paragraph{Agentic LLM systems for clinical genomics.}
Turning such signal into action is where agentic LLM systems come in
\citep{wang2025survey,yao2023react,vallabhaneni2026committee}. Recent systems handle gene
prioritization and diagnostic reasoning \citep{zhao2026deeprare,lee2025lamarrvel}, surface undiagnosed
patients from records \citep{singhal2025aortopathies}, interpret variants from literature evidence
\citep{li2025autopm3,depaoli2024varchat,twede2026evidence} or curated databases \citep{lu2025boosting},
and reanalyze stored genomic data as evidence accrues \citep{welland2026reanalysis}; all operate inside
the health system, on structured inputs for an already-seen patient. Deployed pipelines in other domains
triage unstructured text to drive follow-up, such as routing food-rescue feedback
\citep{raman2026rescuelens} or prioritizing health-outreach contacts \citep{mate2022rmab}, but not user-authored clinical text.

\paragraph{Evaluating clinical LLM pipelines.}
Health LLM work is validated by agreement
against expert annotation \citep{consoli2026sdohgpt,guevara2024sdoh,guo2024llmhealth} or by end-to-end
task success such as trial matching \citep{jin2024trialmatch}; clinical-genetics benchmarks probe
variant interpretation \citep{queen2025cgbench,lin2025benchmarking}, and human-evaluation frameworks
score answers for factuality, harm, and bias \citep{singhal2023clinical}, and clinician panels have judged chatbot replies to patient questions on public forums \citep{ayers2023comparing}. Benchmark scores alone are
increasingly seen as necessary but not sufficient for deployment \citep{raman2026benchmarks}, and the
AI-for-social-impact community urges strengthening evaluation of systems before deployment
\citep{tambe2025nextwave,wilder2025fostering,perrault2020ai4si}. These score a label's accuracy or a clinician-facing
task, not a pipeline whose output is a message drafted for the poster who authored the text.

These four lines each supply part of the picture but not their intersection: to our knowledge, PRISM
is the first to connect population-scale social-media signal to a specific clinical-genetics action, variant
reinterpretation, finding the need, drafting the message, and
checking whether it is fit to send, without contacting anyone.

\section{The PRISM Corpus}
Measuring where reinterpretation need surfaces calls for a corpus that reaches both
the communities where genetics is the subject and the far larger ones where it is an
occasional aside. We assemble PRISM, a corpus of \CorpusPosts{} public posts from
\CorpusSubs{} health-related Reddit communities (subreddits), collected as fully as the public
archive permits, spanning \CorpusYearStart{} to \CorpusYearEnd{}. We pull each community's full
available history rather than a recency-capped API window, so the corpus reflects how the phenomenon grows over time
instead of a snapshot biased toward recent activity.

\paragraph{Communities.}
The \CorpusSubs{} communities are selected to cover the full range of where genetic results are
discussed, grouped by hand into four categories by primary clinical focus:
\textbf{specialist genetics}, where a reinterpretation question is on-topic (hereditary cancer,
clinical genetics and counseling, rare-disease and prenatal testing); \textbf{direct-to-consumer
testing}, built around consumer kits and raw-data interpretation; \textbf{chronic or complex
illness}, conditions with a genetic component; and \textbf{general medical advice}, high-volume
ask-a-doctor forums. Together
these span the concentrated and diffuse ends of where genetics discussion occurs
(Figure~\ref{fig:plane}); the community list and full selection procedure are in the supplement.

\paragraph{Scale and activity.}
The corpus records \CorpusAuthorsApprox{} distinct health-help-seeking authors writing from
\CorpusYearStart{} to \CorpusYearEnd{}, in active discussion rather than dormant archives:
\CorpusReplyPct\% of posts receive at least one reply.

\paragraph{Data and Privacy.}
The corpus is public Reddit data from the Arctic Shift
archive\footnote{\url{https://arctic-shift.photon-reddit.com/}}
\citep{he2026geopolitical,sehgal2026semaglutide}, which honors user removal requests. Per Reddit's
Terms of Use we do not redistribute raw post text; \ReleaseVerb{} the collection, triage, and analysis
code and the method to reconstruct the corpus from the public archive. We keep no direct identifiers:
usernames are replaced by a one-way salted hash at ingest and are not retained raw in the analysis
database or any released artifact, and post text, which may carry indirect identifiers, is not redistributed. Posts are processed by a hosted LLM API
(\TriageModel{} via OpenAI), whose inputs are not used for model training under the provider's
policy;\footnote{\url{https://openai.com/policies/api-data-usage-policies}} the Step-3 Coach
(\texttt{gpt-4o}, OpenAI) and the comparisons in Section~\ref{sec:comparisons} (Claude Code and an
automated judge, Anthropic; Codex, OpenAI) also received post text, never usernames.
Because the study uses only public data, no institutional or clinical records, contacts no individual,
and reports only aggregate results, we had judged it not human-subjects research; such Reddit analyses
are often judged to need no IRB review \citep{proferes2021studying,sehgal2026semaglutide}. We have
requested an IRB determination. Our stakeholder survey on outreach is IRB-approved, and future outreach
needs separate IRB approval.

\section{Method: The PRISM Pipeline}
\subsection{Overview}
PRISM runs three LLM agents with a deterministic ClinVar-resolution step, plus an \emph{Outreach
Coach} that tunes the outreach prompt beforehand and a clinician who reviews the drafts; neither
runs on live posts. Each stage appends to a SQLite table the next reads, tagged with model, prompt version, and run, so
every result traces to a specific configuration. Figure~\ref{fig:overview} summarizes the pipeline; we detail
each agent next.

\subsection{The Triage Agent}
The triage agent is the only one that sees the whole corpus, so it is built to be
cheap and decisive: it reads a post and returns one of three labels in a
single call, with no tools or retrieval. The classes are
\emph{testing-warranted} (B1); \emph{reinterpretation-relevant} (B2), where the author
references a known result whose interpretation may have changed
\citep{landrum2018clinvar,richards2015acmg,clingen2025resource,preston2022vci}; and \emph{irrelevant}. Given the subreddit, title, and body under a fixed instruction prefix, it returns a structured label with a confidence and a one-sentence rationale.

\subsection{The Branch-2 Agent}
Only reinterpretation-relevant posts reach the Branch-2 agent, so this stage can
afford to be careful, and it is handed only the part of the task that needs
judgment: reading the poster's free-text account and pulling out the genetic
identifiers as a (gene, variant, test-date) tuple. The pipeline then queries
ClinVar \citep{landrum2018clinvar} for the variant's most recent qualifying submission
dated at or before the stated test date (or the post date when no test date is stated) and
again for its current aggregate classification, and labels the change
(reclassified, unchanged, conflicting, or not found). A variant that resolves to a unique ClinVar
record whose most recent qualifying submission as of that date differs from the current aggregate is a
\emph{candidate classification difference}: \emph{candidate} because the difference is detected in the
ClinVar record programmatically, not adjudicated as a genuine reclassification, so outreach only
suggests the result may warrant review. PRISM consumes ClinVar as returned and
does not ask the model to judge pathogenicity, a task on which large language models still overclassify and answer
inconsistently \citep{lin2025benchmarking,queen2025cgbench}, and on which deep-learning predictors diverge from clinical-grade classification \citep{kong2025discordance}.

\subsection{The Outreach Agent}
For a candidate classification difference, the outreach agent drafts the message
a knowledgeable peer might send rather than one a clinician would: it names the
poster's gene, states in plain language what the classification was and what it is
now, and points them to a genetic counselor for an up-to-date interpretation, without
offering a diagnosis. Every draft closes with a fixed disclaimer
identifying it as automated research output, and the agent runs in a dry-run: drafts are
written to the database but never sent; clinician review of LLM-drafted patient messages is being
piloted in clinical settings \citep{garcia2024draft}. The peer register describes tone only: messages
are never posted, and any deployment would disclose up front that a message is automated.

\subsection{The Outreach Coach Agent}
The quality of those drafts turns on the outreach prompt, which is easy to over-fit when
edited by hand. We therefore delegate the prompt to a fourth agent, the Outreach Coach, which
edits the instructions rather than any message: each round it receives the current prompt, the
rubric, and a batch of reviewer-scored drafts, and returns a revised prompt, an instance of automatic prompt optimization from reviewer feedback \citep{pryzant2023apo}. It is judged only by
how its next batch is rated, and turns each correction into a general rule rather
than a patch. An unchanged prompt signals convergence; while a failure remains open, the Coach is nudged and
retried, up to three times. Its refinement trajectory, including how each step stopped, is reported in
the experiments.

\section{Experiments and Results}\label{sec:experiments}
PRISM is evaluated in two settings. First, a
synthetic \emph{control set} with known answer keys lets us develop the pipeline
and confirm that it behaves correctly before the corpus-of-record run. Second, we
apply the pipeline to real data: we measure the reinterpretation landscape over the
full corpus and evaluate outreach quality through a four-step protocol that moves
from internal reviewers to expert clinical review. Step~1 runs in the control setting;
Steps~2 to~4 run in the real setting. We report each experiment below.

\subsection{Experimental Setup}
\textbf{Data.} PRISM is measured on the PRISM corpus (\CorpusPostsApprox{} posts) and
validated on a synthetic \emph{control set} of \CtrlSetTotal{} posts with known answer
keys, built backward from real ClinVar variants (construction in the supplement).

\textbf{Models and configuration.} All four agents call \TriageModel{} through the
OpenAI Agents SDK; the one exception is the Step-3 Coach run, a development tool, which used
\texttt{gpt-4o}. Every table and figure is regenerated from the run database by a single script.

\textbf{Metrics.} A \emph{class rate} is the fraction of posts with a given class in the set considered, an exact census over the full corpus. Triage and Branch-2 are
scored against the control answer keys; outreach drafts are scored on a predefined
safety-and-quality rubric (full rubric in the supplement). The dual-reviewer step reports
inter-reviewer agreement on that rubric; the held-out clinician review reports the
reclassification-statement rate (factual dimension) and the holistic acceptability rate
(\% fit to send) as counts over the evaluation set.

\subsection{Setting 1: Control set}
The control set contains \CtrlSetTotal{} synthetic posts with known answer keys:
\CtrlBTwoSet{} reinterpretation-relevant (B2), \CtrlBOneSet{} testing-warranted
(B1), and \CtrlIrrSet{} irrelevant. This paper concerns Branch-2, so the B1
and irrelevant posts serve as \emph{negative controls} that a correct system must
keep out of it. Each B2 post is built backward from a real ClinVar gene, variant, and
classification history, so the correct triage label and reclassification outcome are fixed in
advance. Each variant is rendered into one Reddit-style post that carries only the identifier for its
tier: an rsID; a gene with a c.\ or p.\ change; or a gene alone, which a correct pipeline must
decline to resolve. Posts are generated by \TriageModel{} with a fixed seed, near-duplicates are
removed by sentence-embedding similarity, and each post carries a date-bracketed answer key. We run the pipeline on the control set end to end and check each stage against the
known answers before the corpus-of-record run.

\begin{figure*}[!t]
\centering
\includegraphics[width=0.95\textwidth]{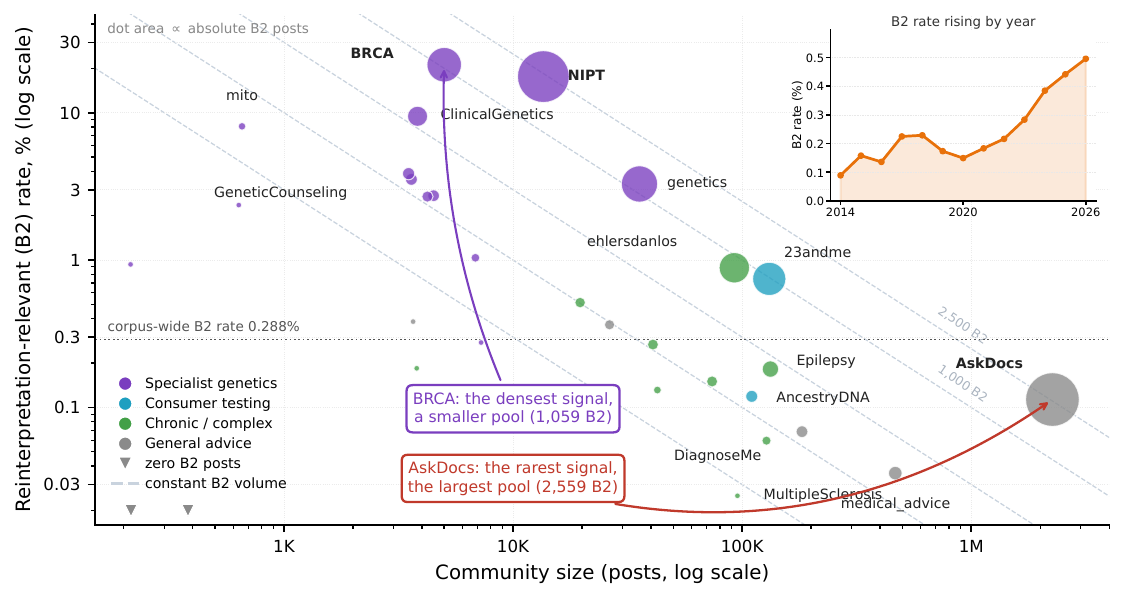}
\caption{\textbf{Rate and volume dissociate.} Each of the \CorpusSubs{} communities is
positioned by size (posts) against reinterpretation-relevant (B2) rate, both on log scales;
point area scales with B2 count, dashed
contours mark constant B2 volume, and the dotted line is the corpus-wide rate (\BTwoPct\%).
Specialist genetics communities (purple) sit upper-left (small, high-rate) and general-advice forums
(grey) lower-right (large, low-rate). \emph{Inset:} the B2 rate by year (2026 is a partial year).}
\label{fig:plane}
\end{figure*}

\textbf{Triage.} The model recovers \CtrlBTwoRecall\% of the true
B2 posts (\CtrlBTwoOK{} of \CtrlBTwoTrue{}) and assigns B2 with a precision of
\CtrlBTwoPrec\% (\CtrlBTwoOK{} of \CtrlBTwoRouted{}). Only \CtrlOverCalls{} of the
\CtrlNegTotal{} negative-control posts are misrouted into B2, so the B2 class is
well separated from the two non-B2 classes.

\textbf{Branch-2.} Variant resolution is graded
against the same answer keys. The control B2 posts are stratified into \emph{identifier tiers}
by how much locating information they carry: two tiers carry a resolvable identifier and a third
gives a gene name alone (definitions and counts in the supplement). Over the two resolvable
tiers, extraction passes \SOneExtCorrect{} of \SOneExtTotal{} posts to ClinVar (\SOneExtPct\%) and
resolution recovers the right variant in \SOneResCorrect{} of \SOneResTotal{}
uniquely-mapping cases (\SOneResPct\%); on the gene-only tier, all \SOneGeneOnlyN{} posts are
correctly declined, none forced to a specific variant. The per-tier, stage-by-stage breakdown is in the supplement.

\textbf{Step~1 outreach.} As a control-setting sanity check, a single author-reviewer scored
\CtrlReviewBTwo{} control-B2 drafts against a four-dimension rubric; all passed.

\subsection{Setting 2: Real data}
\subsubsection{Triage: the reinterpretation landscape}
\paragraph{Rare in aggregate.}
Over the full corpus, reinterpretation-relevant posts are rare:
\BTwoCount{} of the \TotalTriaged{} triaged posts (\BTwoPct\%). This \BTwoPct\% is an
agent-predicted rate, not a measured prevalence: the triage labels are validated on the control
set (\CtrlBTwoPrec\% B2 precision) and, for precision, on real posts (below). Per-community (per-subreddit) rates are listed in the
supplement.

\paragraph{Precision on real posts.}
Two authors labeled \PilotN{} pilot posts (supplement), blind to model labels
($\kappa = \PilotKappa{}$); on the \PilotAgreeN{} they agreed on, triage B2 precision is
\PilotBTwoPrecision\% (\PilotBTwoTP{} of \PilotBTwoPredN{}), and \PilotSpecPrecision\%
(\PilotSpecTP{} of \PilotSpecPredN{}) in specialist genetics; elsewhere B2 posts were too few to
estimate, but false positives there would lower the \GenRate\%. Because the pilot was sampled by
an earlier triage run's predicted labels, this is precision, not recall: at \BTwoPct\% B2, \RecallSampleN{} random posts hold
about \RecallExpectedBTwo{}, and even \RecallExpectedBTwo{} of \RecallExpectedBTwo{} gives a 95\%
recall interval of \RecallCILo\% to 100\% \citep{webber2013recall}.

\paragraph{Concentrated in specialist communities.}
The aggregate rate understates where reinterpretation need concentrates
(Figure~\ref{fig:plane}).
Reinterpretation-relevant discussion concentrates in specialist genetics
communities, which comprise \SpecCorpusPct\% of the corpus but contain
\SpecBTwoShare\% of all reinterpretation-relevant posts at a rate of \SpecRate\%;
general medical-advice communities comprise
\GenCorpusPct\% of the corpus at a rate of \GenRate\% (the community-level rate spans roughly
\RateSpread$\times$; Figure~\ref{fig:plane}). The \BTwoPct\%
aggregate rate reflects corpus composition: rate-dense in specialist communities, numerically
substantial in high-volume general-advice ones. Because the subreddit name is part of
the triage input, we reran triage without it on a stratified sample from each community category
(content-only ablation): the contrast held, with an estimated \AblSpecRate\% in specialist genetics
versus \AblGenAdvRate\% in general medical advice. A reasoning audit of the triage rationales is
reported in the supplement.

\paragraph{Cost and scale.} Screening the full corpus is inexpensive: a single clean end-to-end run
costs about \$\CostCleanTotalUSD{} in API fees (\$\CostPerPost{} per post, measured against the
OpenAI usage export), with triage over all \CorpusPostsApprox{} posts dominating and the two
downstream stages adding under half a percent (per-stage breakdown in the supplement). The
computational cost of a population-scale reinterpretation screen is therefore within reach of a single
laboratory or registry, though operational, governance, and clinical-review costs are separate; the
supplement also estimates the run's energy and carbon footprint from published inference-energy
estimates \citep{oviedo2026energy}.

\paragraph{Growing over time.}
The annual share of posts assigned B2 rose roughly fivefold since 2014 (Figure~\ref{fig:plane},
inset), a rise that holds within general medical-advice communities as well as overall and so is not
solely a composition effect.

\subsubsection{Branch-2: the outreach pool}
\paragraph{Control spike.}
As an execution check, \SpikeControlN{} known-answer control posts injected into the real B2 stream
and processed in one interleaved pass matched the answer keys at extraction \SpikeExtPct\% and
resolution \SpikeResPct\% over the two resolvable tiers, with all \SpikeGeneOnlyN{} gene-only posts correctly declined
(per-tier breakdown in the supplement).

\paragraph{Real pool.}
Applying the same Branch-2 pass to the \BTwoCount{} real reinterpretation-relevant posts
yields the outreach pool. Most posts are set aside before any
ClinVar comparison: \BTwoNotActionable{} are not actionable under Branch-2, carrying no resolvable
uncertain result (we do not adjudicate them as false positives), and \BTwoNotResolvable{} name a variant that cannot be resolved in
ClinVar. \BTwoResolvedCV{} resolve to a ClinVar classification; of these,
\BTwoUnchanged{} are unchanged as of the anchor date and \BTwoOutreach{} changed
(the remainder have no datable ClinVar history or a non-standard significance label). The
\BTwoOutreach{} candidate classification differences are the outreach pool
(Table~\ref{tab:b2outreach} splits them by reclassification type): most are a move to
conflicting classifications (\OutToConflicting{}), with \OutToPathogenic{} shifting from an uncertain
submission at the test date to pathogenic or likely pathogenic, the potentially highest-consequence change. The full breakdown by ClinVar
classification is in the supplement. For this \BTwoOutreach{}-post pool, outreach message
generation and quality evaluation follow in Steps~2 to~4 (below).

\paragraph{Dated split.}
As run, the pipeline compares one past submission with ClinVar's current overall classification, so a
difference can also reflect disagreement between labs that already existed at the test date. To
separate the two, we checked each of the \BTwoOutreach{} against ClinVar's overall classification in
\ArchReleases{} archived monthly releases (2017 to 2026), compared with PRISM's July 10, 2026 lookup:
\ArchDated{} of the \BTwoOutreach{} (\ArchDatedPct\%) changed or first appeared in ClinVar after the
test date, and \ArchUnchanged{} (\ArchUnchangedPct\%) were unchanged, including \ArchConflUnchanged{}
of the \OutToConflicting{} transitions to conflicting. We report this dated split alongside the
original \BTwoOutreach{}, and we have implemented the archived-release lookup in PRISM. The clinician
review (Step~4) checks drafts against the pipeline's classifications, not archived releases.

\begin{table}[t]
\centering \footnotesize
\setlength{\tabcolsep}{6pt}\renewcommand{\arraystretch}{0.95}
\begin{tabular}{llrrr}
\toprule
\textbf{From} & \textbf{To} & \textbf{Pool} & \textbf{Dev} & \textbf{Test} \\
\midrule
VUS & Conflicting & 53 & 8 & 12 \\
P/LP & Conflicting & 40 & 11 & 8 \\
B/LB & Conflicting & 19 & 6 & 4 \\
\rowcolor{gray!15} VUS & P/LP & 5 & 2 & 3 \\
Conflicting & B/LB & 2 & 1 & 1 \\
B/LB & VUS & 1 & 0 & 1 \\
P/LP & B/LB & 1 & 0 & 1 \\
P/LP & VUS & 1 & 1 & 0 \\
VUS & B/LB & 1 & 1 & 0 \\
\midrule
\textbf{Total} & & \textbf{123} & \textbf{30} & \textbf{30} \\
\bottomrule
\end{tabular}

\caption{The \BTwoOutreach{}-post outreach pool by reclassification type (historical vs current
ClinVar classification). VUS $=$ uncertain, P/LP $=$ pathogenic or likely pathogenic, B/LB $=$ benign
or likely benign; the VUS-to-P/LP row (potentially highest-consequence) is shaded. Dev and Test are the disjoint
prompt-optimization and clinician-review subsets.}
\label{tab:b2outreach}
\end{table}

\newcommand{\prismstep}[2]{\faAngleRight~\texttt{#1}: #2\\}
\begin{figure*}[!t]
\centering
\begin{minipage}[t]{0.49\textwidth}
\begin{tcolorbox}[colback=white, colframe=black!80, coltitle=white, fonttitle=\bfseries\footnotesize,
  title={\textcolor{green!65!cyan}{\faDna}~Test-set example \textbullet\ uncertain $\rightarrow$ pathogenic}, boxrule=0.8pt, arc=2pt,
  left=5pt, right=5pt, top=3pt, bottom=3pt]
\footnotesize
{\bfseries\faComments~The post} \textit{(paraphrased)}\\[1pt]
\colorbox{blue!8}{\parbox{\dimexpr\linewidth-2\fboxsep\relax}{\itshape ``I ran my raw DNA data and found a BRCA2 variant listed as uncertain. Should I worry?''}}
\tcbline
{\bfseries\faCogs~What PRISM does}\\[1pt]
\prismstep{resolve}{\textbf{a \emph{BRCA2} frameshift variant}}
\prismstep{ClinVar}{as of the post \textbf{Uncertain} $\rightarrow$ now \colorbox{red!72!black}{\textcolor{white}{\bfseries\,Pathogenic\,}}}
\faAngleRight~\texttt{draft}: a disclaimer-bearing outreach message
\tcbline
{\bfseries\faRobot~PRISM's draft} \textit{(excerpt)}\\[2pt]
\colorbox{green!14}{\parbox{\dimexpr\linewidth-2\fboxsep\relax}{\itshape ``\ldots the BRCA2 variant you found was uncertain when you posted, but ClinVar now classifies it as pathogenic. ClinVar is a free NIH database that compiles classifications from many labs, so this reflects more accumulated evidence\ldots please consider following up with a genetic counselor for an up-to-date interpretation.'' [disclaimer]}}
\tcbline
{\bfseries\faStethoscope~Clinician review}~~\colorbox{green!50!black}{\textcolor{white}{\bfseries\,\faCheckCircle\ Fit to send\,}}
\end{tcolorbox}
\end{minipage}\hfill
\begin{minipage}[t]{0.49\textwidth}
\begin{tcolorbox}[colback=white, colframe=black!80, coltitle=white, fonttitle=\bfseries\footnotesize,
  title={\textcolor{green!65!cyan}{\faDna}~Test-set example \textbullet\ likely pathogenic $\rightarrow$ conflicting}, boxrule=0.8pt, arc=2pt,
  left=5pt, right=5pt, top=3pt, bottom=3pt]
\footnotesize
{\bfseries\faComments~The post} \textit{(paraphrased)}\\[1pt]
\colorbox{blue!8}{\parbox{\dimexpr\linewidth-2\fboxsep\relax}{\itshape ``My report calls my MYH7 variant uncertain, but ClinVar had it as likely pathogenic. Why?''}}
\tcbline
{\bfseries\faCogs~What PRISM does}\\[1pt]
\prismstep{resolve}{\textbf{an \emph{MYH7} missense variant}}
\prismstep{ClinVar}{as of post \textbf{Likely pathogenic} $\rightarrow$ \colorbox{orange!85!black}{\textcolor{white}{\bfseries\,Conflicting\,}}}
\faAngleRight~\texttt{draft}: a disclaimer-bearing outreach message
\tcbline
{\bfseries\faRobot~PRISM's draft} \textit{(excerpt)}\\[2pt]
\colorbox{red!10}{\parbox{\dimexpr\linewidth-2\fboxsep\relax}{\itshape ``\ldots there is not enough evidence to determine whether the variant is harmful or harmless, and it usually does not affect medical decisions\ldots''}}
\tcbline
{\bfseries\faStethoscope~Clinician review}~~\colorbox{red!62!black}{\textcolor{white}{\bfseries\,\faTimesCircle\ Unfit to send\,}}\\[2pt]
overreach: ``harmful or harmless'' should be ``associated with disease''; harm includes the psychological effect of carrying it.
\end{tcolorbox}
\end{minipage}
\caption{\textbf{Held-out examples: PRISM's recontact drafts and the clinician's verdict.} Two of the $N{=}\OutreachClinicianN{}$ test-set drafts, one fit, one unfit, paraphrased and excerpted for privacy/Terms-of-Use. Across the held-out set, \StepFourFactualN{} of \StepFourFitTotal{} stated the reclassification correctly, while \StepFourFitN{} of \StepFourFitTotal{} drafts (\StepFourFitPct\%) were rated fit to send under the clinician's overall call.}
\label{fig:outreach_examples}
\end{figure*}

\subsubsection{Outreach: evaluating the messages}
\paragraph{Steps~2--3: prompt development.}
The outreach agent drafts a message for each of the \BTwoOutreach{} pool cases; from these we sample
a development set of \OutreachDevN{}, stratified across the four framing branches. In Step~2, a single
reviewer scores each draft against the four-dimension rubric while the Outreach Coach revises the
generator prompt and the agent redrafts; the rubric-pass rate rose from \OutreachDevStart\% to
\OutreachDevEnd\% over \OutreachNVersions{} versions. That rubric was derived from the synthetic
control set, where a message trivially fits a post we generated ourselves; on real posts it proved too
lenient~\citep{zhuang2020misaligned}, missing whether a message fits the poster's specific context. A senior author
therefore reviewed the real drafts and expanded the rubric to six dimensions, adding the contextual
and explanatory axes; we applied that feedback by hand to both the generator and Coach prompts as the
Step-3 starting point. Step~3 reruns the loop under the stricter six-dimension rubric with two
independent reviewers: inter-reviewer agreement reached
\StepThreeAgreePct\% (\StepThreeAgreeCILo--\StepThreeAgreeCIHi\% Wilson interval), up from
\StepThreeAgreeStartPct\%, the \StepThreeSplitN{} remaining split decisions all on
\StepThreeSplitDim{}. The Coach, run here on \texttt{gpt-4o} as a development tool, returned the prompt unchanged after
one round, and we froze it without the nudged retries of the stopping rule. With the prompt
\emph{frozen}, the held-out clinician review in Step~4 judges whether its messages are fit to send. Full trajectories and
the frozen prompt are in the supplement.

\paragraph{Step~4: outreach generator and clinician.}
Step~4 tests whether the frozen pipeline generalizes to messages it was never tuned on. The clinician
set is drawn entirely from the \OutreachHeldoutN{} outreach cases not used in development, so it is
disjoint from the \OutreachDevN{} development cases; no tuned-on message is rated. From this held-out
pool we take \OutreachClinicianN{} messages: every rare reclassification available in the held-out
pool (a move to pathogenic, benign, or uncertain, the potentially highest-consequence but scarce changes) is
included, and the remainder is filled by sampling from the abundant conflicting cases. A clinician
not involved in tuning or development rates each held-out
message against the six-dimension rubric and makes an overall fit-to-send call. For each draft the
clinician sees the post, the pipeline's gene, variant, and ClinVar transition, and the draft, and
scores six binary dimensions (factual, non-coercive, non-alarming, no overreach, explanatory,
contextualized); the rubric prespecifies
that this call is a holistic judgment that may fail a message for a documented safety reason
beyond the six dimensions, following the factuality,
harm, and appropriateness axes for medical LLM outputs \citep{singhal2023clinical}
and the unsafe-advice bar general chatbots fail \citep{draelos2026unsafe}; supportive-message quality has been formalized similarly for emotional-support dialog \citep{liu2021esconv}. Because every
rated message is held out from optimization, the factual-grounding and acceptability rates are the
paper's primary outreach result. All \StepFourFitTotal{} posts were rubric-scored and none was
flagged as inappropriate for Branch-2 (\StepFourFitTotal{} of \StepFourFitTotal{} appropriately
routed); \StepFourFactualN{} of \StepFourFitTotal{} drafts stated the reclassification correctly,
\StepFourAllSixN{} met all six rubric dimensions, and \StepFourFitN{} of \StepFourFitTotal{}
(\StepFourFitPct\%) were fit to send. Because the set was enriched for the rare, highest-consequence
transitions, these rates describe the \StepFourFitTotal-message evaluation set rather than the full
\BTwoOutreach-candidate pool. Figure~\ref{fig:outreach_examples} illustrates these with worked
examples and their verdicts. A multi-clinician evaluation will follow our ongoing stakeholder survey.

\subsection{Additional comparisons}\label{sec:comparisons}
\textbf{Web-search agents.} We gave PRISM's prompts to two web-search agents, Claude Code (Claude
Sonnet 4.6) and Codex (\texttt{gpt-6-astra}). On the \PilotAgreeN{} labeled pilot posts, their B2
precision was higher (\EOneClaudePrecision\% and \EOneCodexPrecision\% vs \PilotBTwoPrecision\%) at
far higher cost (USD \EOneClaudeCost{} and \EOneCodexCost{} vs \PrismCostPerBTwoPost{} per post for triage plus Branch-2,
\EOneCostRatioLo{} to \EOneCostRatioHi{} times). We use \TriageModel{} because larger models cost more
\citep{ong2025routellm}, and tuned prompts can make small models competitive \citep{khattab2024dspy}.
On the same variants, PRISM's lookup returned the correct current ClinVar classification for
\EOnePrismNowOnClaude{} of \EOneClaudeNowN{} and \EOnePrismNowOnCodex{} of \EOneCodexNowN{}, the agents
for \EOneClaudeNowAgree{} of \EOneClaudeNowN{} and \EOneCodexNowAgree{} of \EOneCodexNowN{} (Codex
abstained on \EOneCodexNowAbstain{}). For the test-date classification the agents did better (Claude
\EOneClaudeThenAgree{} vs PRISM \EOnePrismThenOnClaude{} of \EOneClaudeThenN{}), as PRISM uses one past
submission; the archived-release lookup addresses this.

\textbf{Outreach.} On the \ETwoN{} held-out cases, with one writing model (\TriageModel{}) and an
automated judge (Claude Sonnet 4.6, six-dimension rubric), a fixed template was judged fit to send in
\ETwoTemplateFit{} of \ETwoN{} cases and our reported prompt in \ETwoReportedFit{}; direct extraction
and lookup is already PRISM's design. In a Coach ablation, fit to send was \ETwoNoCoachFit{} vs
\ETwoCoachFit{} of \ETwoN{} with the initial prompt (v0) vs the Coach-tuned prompt (v5) (McNemar $p = \ETwoCoachP{}$), so its held-out
contribution is not established, though in development it raised acceptance from \OutreachDevStartN{}
to \OutreachDevEndN{} of \OutreachDevN{}. On \OutreachClinicianN{} clinician-rated drafts, the judge
matched our clinician no better than chance ($\kappa = \ETwoJudgeKappa{}$), so these are automated
checks, not quality measures.

\section{Discussion}
PRISM surfaces \BTwoAuthors{} distinct hashed author identifiers across the \BTwoCount{} reinterpretation-relevant
posts, each having publicly raised a genetic result whose interpretation may have changed.
Because these posts are public, they are observable, unlike the far larger population who never post.
PRISM maps where
publicly expressed reinterpretation signals concentrate, not the prevalence of
clinically confirmed unmet need; confirming that any individual requires
reinterpretation would need prospective, consented clinical evaluation.

The concentration of this discussion (Figure~\ref{fig:plane}) indicates where a
recontact effort could focus, a signal the diffuse-responsibility model does not provide.
This reframes recontact as equitable access: because the signal surfaces in the public spaces where people seek
answers rather than a clinic that may have lost track of them, a system built on it could,
with consent, reach patients no longer in specialist care \citep{muir2022reclass}.

On the held-out set (Section~\ref{sec:experiments}), the outreach result separates factual pipeline
performance from communication acceptability. The \StepFourFactualN{} factually grounded messages are
a clinician-checked, positive-path factual validation of the pipeline on real posts: each names only
a gene or variant the post supports and states the pipeline-resolved transition accurately, so
extraction, resolution, and message generation held together across these cases. Given the pipeline's resolution, the main remaining challenge was therefore not variant
identification but communication: \StepFourFactualUnfitN{} of the \StepFourFactualN{} factually
grounded drafts were still unfit, the failures being inconsistent referral (a genetic counselor
versus a physician), unnecessary repetition, or statements about what a provider might do. Of the messages that
met all six dimensions, only \StepFourAllSixUnfitN{} was withheld, on holistic grounds alone, a
language-level concern the six dimensions do not encode; this is why fit-to-send is a holistic judgment
rather than a conjunction of the dimensions. PRISM is therefore intended for clinician-in-the-loop use rather than autonomous outreach: given the pipeline's
resolution on these cases, the evaluation isolates communication and context as the specific barriers that
remain, which is exactly where clinician review is positioned to help.

\paragraph{Patterns from clinical review.}
Clinician review also flagged two patterns in the agent's drafts that now guide prompt revision, as
intended by our iterative clinical evaluation. First, \HarmlessN{} of the \OutreachClinicianN{}
drafts use the prompt's ``harmful or harmless'' wording with no move to uncertain, each where the
poster said uncertain; the clinician flagged it in \HarmlessFlaggedN{} of them, and we have limited
this wording to moves to uncertain in a revised prompt (the reported results use the frozen prompt). Second, of the \StepFourVusPlpN{} messages on a move from an
uncertain submission to pathogenic or likely pathogenic, \StepFourVusPlpFitN{} was fit to send; the
other \StepFourVusPlpNoGeneN{} named a variant but not its gene, and the clinician preferred referral
to the ordering provider.

\paragraph{A template for unmet health needs.}
PRISM is a template for finding unmet health needs online, beyond reclassification: the triage
criteria define the need, the extraction fields its identifiers, the reference database the evidence,
and the outreach prompt the message. Replacing ClinVar with CPIC drug-gene guidelines, for example,
could identify pharmacogenomic posts. Setting~2 adds two reusable findings: such needs are rare but
concentrated (\SpecRate\% versus \GenRate\%), and factual accuracy is insufficient
(\StepFourFactualN{} of \OutreachClinicianN{} grounded; \StepFourFitN{} fit to send).

\paragraph{Path to deployment.}
A hospital, laboratory, or registry would host PRISM with a clinician reviewing each candidate;
individuals could run the \ifreleased public code\else code once it is public\fi. PRISM need not run continuously: about \MonthlyPostsApprox{}
posts are added each month and ClinVar issues new releases, so each release triggers a run on new posts
(about USD \MonthlyRunUSD{} per month, versus about USD \FullRerunUSD{} for a full rerun; daily runs
cost the same) and a re-check of stored variants. Our IRB-approved survey of patients, caregivers, and
clinicians on AI-identified outreach is under way and will guide a stakeholder-centered design for
real-world deployment.

\paragraph{Clinical significance.}
Reclassification is common, yet recontact is inconsistent and, when it occurs, falls to clinic staff rather than the laboratory
\citep{makhnoon2023recontact,muir2022reclass}; existing channels reach mostly patients still in care
or in a registry \citep{savatt2021recontact}, and responsibility stays diffuse
\citep{david2019recontact,appelbaum2020duty}. Clinical review indicates that the appropriate unit for this setting is a minimal factual
update, stating that a poster's variant moved between classifications, with a
pointer to follow-up, rather than a full clinical interpretation: such an update is intended to be
low-risk and informative, flagging the change without acting on it. Used this way, PRISM is a surfacing layer that complements clinic- and registry-based
recontact rather than replacing clinical judgment.

\paragraph{Limitations.}
PRISM is a research prototype, and several limitations constrain any move toward live use.
\emph{Deployment.} PRISM runs as a dry run and sends nothing; because general-purpose chatbots
return unsafe advice to patients at a non-trivial rate \citep{draelos2026unsafe}, a patient-facing
system must meet a higher safety standard; community replies (\BTwoReplied\% of
reinterpretation-relevant posts receive at least one) are informal peer support without validated
interpretation~\citep{weitzman2011social}, so the recontact gap is motivation, not a demonstrated outcome; and public
availability does not imply consent, and embedding interventions in social platforms carries risks \citep{dunn2018social}.
\emph{Representativeness and label validity.} The landscape rates, concentration, and fivefold trend
rest on agent-assigned triage labels whose precision is validated on real posts but whose recall is
not, since a feasible random sample holds too few B2 posts to estimate it.
The corpus is a single platform whose users skew younger, more educated
\citep{pew2025socialmedia}, and toward the United States \citep{sehgal2026semaglutide} (unlike consented on-device panels \citep{reeves2020screenome}), so the
landscape reflects Reddit rather than the general public and could widen the access gap if scaled
without complementary outreach; the triage input includes the subreddit name, although the community
contrast held without it; and PRISM reads
self-reported text, cannot confirm a claimed result, counts hashed accounts rather than unique
individuals, and cannot attribute the fivefold rise to growth in confirmed need rather than shifting
platform or testing patterns.
\emph{Outreach evaluation.} Step 4 is one clinician on \OutreachClinicianN{} held-out drafts from a
stratified sample~\citep{gordon2021disagreement}: it tests factual fidelity and acceptability, not safe
communication or benefit, and it does not re-adjudicate the ClinVar transitions; a multi-clinician
evaluation will follow our ongoing stakeholder survey.
\emph{ClinVar and modeling.} Branch-2 treats ClinVar \citep{landrum2018clinvar} as authoritative and
anchors on a most-recent-submission proxy rather than a reconstructed historical aggregate, so
outreach suggests review rather than asserting a change; because the control set is built from
variants that reclassified, specificity against unchanged variants is not measured; and all pipeline
agents use one model; the web-search agents above are the only cross-family comparison.

\section{Conclusion}
PRISM maps agent-predicted variant-reinterpretation signals at population scale, converting
candidate classification differences into clinician-reviewed drafts without contacting
anyone. On \OutreachClinicianN{} held-out real-post cases, \StepFourFactualN{} drafts were factually
grounded and \StepFourFitN{} were fit to send: factual grounding alone did not ensure acceptable communication, so outreach in this setting calls for clinician review rather than autonomous sending. Whether such surfacing translates into clinical action without
widening the access gap it aims to close is the question we leave for user studies with clinicians,
patients, and other stakeholders.

\section*{Use of Generative AI}
All PRISM pipeline agents use \TriageModel{}; the Step-3 Coach run, a development tool, used
\texttt{gpt-4o}. The authors verified all AI-assisted coding, analyses, and language editing.

\section*{Acknowledgments}
This work was supported by the National Human Genome Research Institute of the National Institutes of Health
under award R01HG012655 and by the NIH Office of the Director under award OT2OD037960. The content is solely
the responsibility of the authors and does not necessarily represent the official views of the National
Institutes of Health. We thank Parul Chaudhary for helpful discussions.

\setlength{\bibsep}{2pt plus 0.3ex}  %
\bibliography{references}

\clearpage
\includepdf[pages=-]{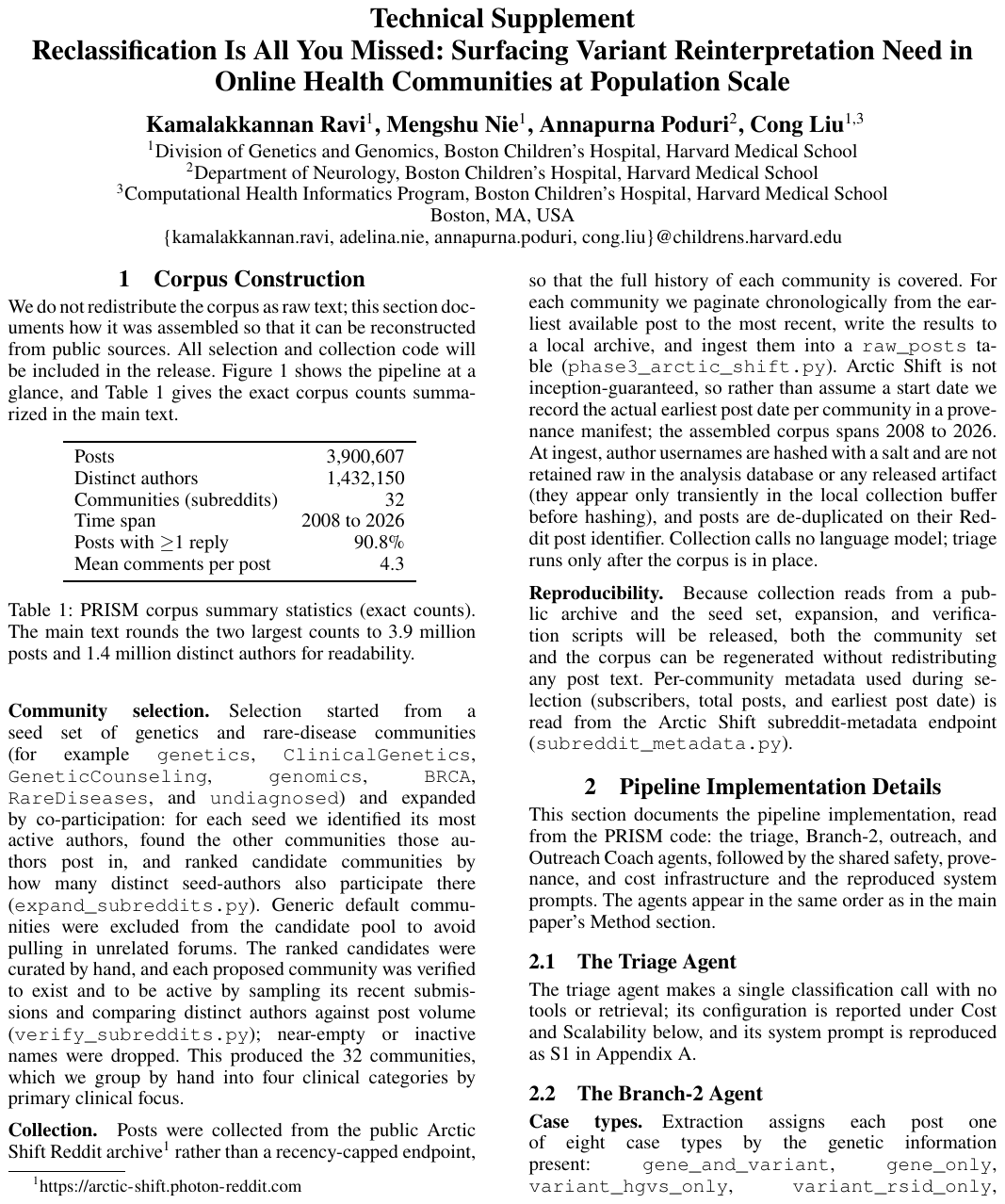}
\end{document}